\documentclass[lettersize,journal]{IEEEtran}
\usepackage{amsmath,amsfonts}
\usepackage{algorithmic}
\usepackage{array}
\usepackage[caption=false,font=normalsize,labelfont=sf,textfont=sf]{subfig}
\usepackage{textcomp}
\usepackage{stfloats}
\usepackage{url}
\usepackage{verbatim}
\usepackage{graphicx}
\usepackage{float}
\usepackage{multirow}
\usepackage{xcolor}
\usepackage[table]{xcolor}
\usepackage{caption}
\usepackage{threeparttable}
\usepackage{makecell}
\def\BibTeX{{\rm B\kern-.05em{\sc i\kern-.025em b}\kern-.08em
    T\kern-.1667em\lower.7ex\hbox{E}\kern-.125emX}}
\usepackage{balance}
\usepackage[colorlinks,
linkcolor=red,
anchorcolor=blue,
citecolor=green
]{hyperref}
\usepackage{cite}

\begin{document}
\title{B2N3D: Progressive Learning from Binary to N-ary Relationships for 3D Object Grounding}
\author{Feng Xiao, Hongbin Xu, Hai Ci, Wenxiong Kang, ~\IEEEmembership{Member, ~IEEE} 
\thanks{Feng Xiao and Wenxiong Kang are with the School of Automation Science and Engineering, South China University of Technology, Guangzhou, China. Hongbin Xu is with ByteDance Seed, Beijing, China. Hai Ci is with Show Lab, National University of Singapore, Singapore.  
Wenxiong Kang is the corresponding author (email: auwxkang@scut.edu.cn).

The code is released at \href{https://github.com/onmyoji-xiao/B2N3D}{{https://github.com/onmyoji-xiao/B2N3D}}.
}}

\markboth{Journal of \LaTeX\ Class Files,~Vol.~18, No.~9, September~2020}%
{How to Use the IEEEtran \LaTeX \ Templates}

\maketitle


\begin{abstract}
Localizing the object with natural language descriptions is essential in 3D scene understanding. The descriptions often involve multiple spatial relationships to distinguish similar objects, making 3D-language alignment difficult.
Current methods only model relationships for pairwise objects, ignoring the global perceptual significance of n-ary combinations in multi-modal relational understanding.
To address this, we propose a novel progressive relational learning framework for 3D object grounding. We extend relational learning from binary to n-ary to identify visual relations that match the referential description globally. Given the absence of specific annotations for referred objects in the training data, we design a grouped supervision loss to facilitate n-ary relational learning. In the scene graph created with n-ary relationships, we use a multi-modal network with hybrid attention mechanisms to further localize the target within the n-ary combinations. 
Experiments and ablation studies on the ReferIt3D and ScanRefer benchmarks demonstrate that our method outperforms the state-of-the-art, and proves the advantages of the n-ary relational perception in 3D localization.
\end{abstract}


\section{Introduction}
\IEEEPARstart{In} recent years, applying natural language to 3D scene understanding has become a hot topic \cite{yan2023comprehensive}. Tasks involving reasoning about real-world scenes, such as 3D dense captioning \cite{mao2023complete} and visual question answering \cite{ye20223d}, have garnered significant attention. 3D object grounding aims at associating textual descriptions with specific objects \cite{chen2020scanrefer, achlioptas2020referit3d}. It requires building a vision-language model to reason about object contextual relationships.

\begin{figure}[!t]
  \centering
  \includegraphics[width=\linewidth]{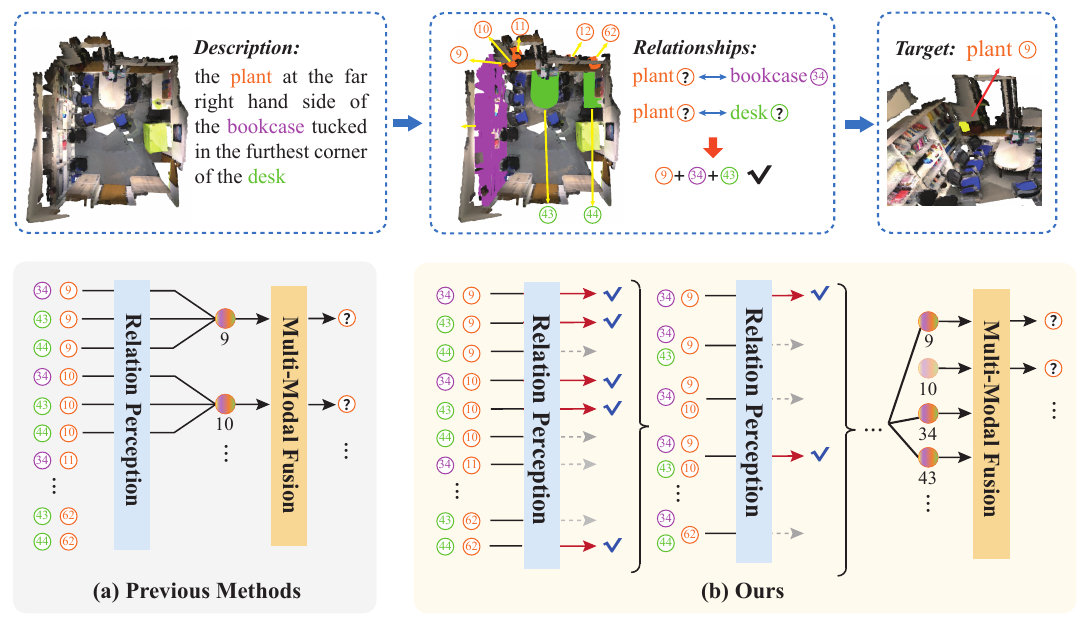}
  \caption{The target or referred objects mentioned in the description often have multiple similar instances present in the scene like ``plant'' and ``desk'' in the figure. Accurately localizing the target needs to identify the correct combination of objects that jointly satisfies the multi-relational constraints (``plant\_9 desk\_43 bookcase\_34''). Compared to the previous methods, which only perceives the relationships of paired objects, our method employs a progressive approach to learn relational composition, achieving global perception of n-ary relationships. }
  \label{fig:start}
  \vspace{-0.6cm} 
\end{figure}

A real-world scene often contains multiple similar objects and has a complex spatial arrangement, people must describe the specific target through referential relationships. As shown in Fig. \ref{fig:start}, understanding the spatial relationships between ``plant-bookcase'' and ``plant-desk'' is necessary for localizing the described target among the five ``plant''. 
Due to the inherent differences between the visual and textual modalities, simultaneously mapping multiple relationships in the text to entities is a significant challenge for 3D object grounding. 
Recent methods mainly use Transformer-based frameworks to achieve 3D-language alignment for localization \cite{he2021transrefer3d, roh2022languagerefer, yang2021sat, zhao20213dvg, huang2022multi, xu2024multi,chang2024mikasa, zhang2024cross}. Some works \cite{yuan2021instancerefer, chen2022language, feng2021free, huang2021text, xiao2025as3d} create scene graphs with entities as nodes and aggregate information from neighboring or text-related nodes to learn entity relationships implicitly. Others \cite{chang2024mikasa, cai20223djcg, zhao20213dvg} integrate independent geometric modules into transformers to enhance spatial relational representation. 
However, when using multiple referential relationships to differentiate the target from similar distractors, current methods still often fail to localize the correct object due to limited relational perception.

Many descriptions, such as the example in Fig. \ref{fig:start}, involve multiple entities and relationships. 
Most existing methods perform relational perception only from pair-wise entities. The multi-modal fusion utilizes all relational information that corresponds with the local description (both ``plant\_9 and desk\_43'' and ``plant\_62 and desk\_44'' fulfill ``the plant tucked in the furthest corner of the desk''). This may increase the uncertainty of the object representation. In cases where multiple entity pairs simultaneously adhere to the described relationships, local reasoning is insufficient for precise localization.

Although graph-based models can represent both local neighborhoods and global contexts \cite{chen2024survey}, the scene graphs in current research establish connections between all spatially adjacent or semantically related entities. This also produces incorrect guidance in cross-modal understanding due to the aggregation of excessive noisy information in graph nodes. Our goal is to infer the n-ary combinations that simultaneously satisfy all referential conditions detailed in the description. This process involves a close relationship between the multi-modal fusion feature and n-ary global representation, while ensuring the removal of interference from noisy information.

In this paper, we propose a novel progressive relational learning framework to improve the cross-modal understanding of multi-relation descriptions in 3D object grounding. Unlike the existing object-to-object modeling methods, we explicitly model binary to n-ary relationships to guide the global relational perception in complex scenes. We creatively use the large language model (LLM) to extract entity relationships from free-form utterances for training. Due to the uncertainty of referred objects, we design a grouped supervision loss function to achieve differentiated prediction of the correlation among n-ary combinations. We construct scene graphs using the entity combinations with the highest response to textual descriptions. To reduce the ambiguity of nodes caused by noise or redundant information, we use a graph neural network based on hybrid attention mechanisms to update nodes. It enhances binary spatial perception using self-attention while integrating multi-modal information from n-ary relationships through graph attention and cross-attention. Experiments confirm the advantages of our proposed progressive relational learning module in 3D localization on public datasets, outperforming other methods.

The main contributions are summarized as follows:
\begin{itemize}
\item{We underscore the importance of global relational perception in complex referential descriptions. We introduce a novel method for 3D object grounding, which utilizes progressive relational learning and attention-driven graph learning.}
\item{Our proposed relational learning module achieves global relational perception by progressively understanding text-guided binary and n-ary relationships. This approach offers an innovative supervised strategy to compensate for the absence of specific referential annotations related to entities.}
\item{The scene graph learning utilizes pre-extracted n-ary groups and a multi-modal fusion network with hybrid attention mechanisms, achieving global perception while minimizing the impact of noise information.}
\end{itemize}

\section{Related Work}

\subsection{3D Object Grounding}
The research on 3D object grounding aims to localize the specific object using referential descriptions in 3D scenes. The current mainstream methods rely on the two benchmarks ScanRefer \cite{chen2020scanrefer} and ReferIt3D \cite{achlioptas2020referit3d} for further research. From a modeling structure perspective, these methods can be classified as two-stage and one-stage \cite{liu2024survey}. The two-stage method first employs a 3D detector to identify object proposals and then utilizes a vision-language perception model to determine the target \cite{chang2024mikasa, huang2022multi, zhang2023multi3drefer}. The one-stage method processes scene point clouds end-to-end and directly outputs the target position \cite{wu2023eda, luo20223d, cai20223djcg}. Describing complex scenes often requires multiple spatial relationships to clarify object positions, making the accurate understanding of these relationships a primary issue. Zhao et al. \cite{zhao20213dvg} and Chang et al. \cite{chang2024mikasa} integrate pre-extracted spatial relationship matrices into the transformer to improve relation awareness. Cai et al. \cite{cai20223djcg} improve geometric relationship expressions through the proposed attribute-relation enhancement module. Huang et al. \cite{huang2021text} construct an instance graph to aggregate neighboring relationship information. Chen et al. \cite{chen2022language} point out that modeling only neighboring objects in the graph ignores the distance expression of relationships and propose a spatial self-attention layer. The state-of-the-art methods \cite{zhang2024cross, xu2024multi} focus more on better visual representations of objects from attributes or semantic labels. Additionally, some works utilize large language models for text analysis or data augmentation to enhance the language comprehension capabilities of models \cite{guo2023viewrefer, chen2023unit3d, jia2024sceneverse, zhan2024mono3dvg}. However, these methods still struggle to make accurate distinctions when faced with multiple objects identical to the target category. In this paper, we explicitly model binary and n-ary relations to facilitate global multi-modal alignment, which enables more accurate localization of objects under complex circumstances.
 
\subsection{Scene Graph in Vision-Language}
Scene graphs provide a high-level understanding of visual scenes, typically representing objects as nodes and relational representations as edges \cite{chen2024survey}. Because of the topological homogeneity of visual and textual modalities in scene graphs, scene graphs are widely employed in vision-language tasks to facilitate cross-modal alignment and joint understanding \cite{chen2022multi, deng2023transvg++, yang2020relationship}. 
Graph neural networks (GNNs) can automatically learn the relationships between nodes when paired with scene graph structures. The introduction of the graph attention network (GAT) \cite{velivckovic2017graph}, which incorporates attention attributes, helps understand complex scenes \cite{tu2023relation, chen2025multi, zhang2024question}. In the field of 3D visual understanding, researchers also construct scene graphs to enhance model understanding of geometric relationships \cite{dhamo2021graph, armeni20193d, yan2023comprehensive}. Koch et al. \cite{koch2024open3dsg} use the objects detected by the 3D detector as graph nodes and construct scene graphs with LLM filtering object-pair relationships as edges. Chen et al. \cite{chen2024clip} predict the relationship score by concatenating the features of two objects and inputting them into a fully connected network, which is used to construct edges. Strader et al. \cite{strader2024indoor} utilize logical rules in spatial ontology as prior knowledge, employing LLM to extract common spatial concepts and establish a scene graph. However, the above methods only focus on the spatial or semantic relationships between pairwise objects when constructing scene graphs. They cannot effectively adapt to the 3D localization task with multiple referential relationships. Our method predicts n-ary relationships globally to establish graph edges and enhance vision-language alignment.

\section{Methods}

\subsection{Overview}
\begin{figure*}[t]
  \centering
  \includegraphics[width=0.9\linewidth]{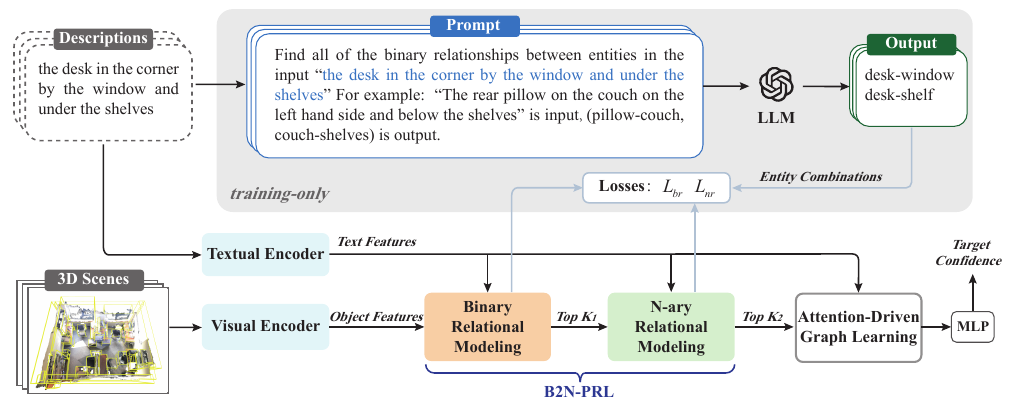}
  \caption{The overall structure of our method \textbf{B2N3D}. It mainly consists of the textual and visual encoders, the B2N-PRL module, and the attention-driven graph learning module. We encapsulate textual descriptions as prompt inputs to the LLM and obtain entity combinations for training. Based on the uncertainty of referred objects (only entity names are known), we design losses $L_{br}$ and $L_{nr}$ to supervise the two relationship modeling processes in the B2N-PRL module. A scene graph is established from the predicted n-ary relationships, and multi-modal fusion is achieved through a graph-structured network. The object with the highest confidence is finally output as the predicted target.}
  \label{fig:overall}
\vspace{-0.4cm}
\end{figure*}

The overall structure of our method is shown in Fig. \ref{fig:overall}. Descriptions and objects within 3D scenes are encoded into text and object features, respectively. Subsequently, the binary-to-n-ary progressive relational learning module (B2N-PRL) module anticipates entity relationships via a text-guided binary to n-ary modelling process. During the training phase, we employ the LLM to extract entity combinations from descriptions related to the referential dependencies, which are used to supervise the progressive relational learning. 
Most referential relationships involve no more than four entities. Therefore, our relational reasoning iterates only twice, with each n-ary combination comprising 2 to 4 objects. The top $K_2$ n-ary combinations with the highest responses are used to construct the scene graphs. The attention-driven graph learning module updates node features and localizes the target by confidence from a multilayer perceptron (MLP).

\subsection{Object-Level Representation}
Since entities are the basic units of relational learning, we prioritize object-level visual feature encoding within scene contexts. Following the two-stage detect-then-match approaches, the scene point cloud is decoded into instance-level point clouds via segmentation networks, then extracts object features using a 3D point cloud encoder. Inspired by \cite{xiao2025as3d}, our method employs a training-free segmentation model based on multiple views to generate class-agnostic object proposals. Simultaneously, we harness projection data during the segmentation process to derive 2D features from the pre-trained multi-modal model. The incorporation of pre-trained 2D visual features addresses the inadequate representation of surface color and texture details in point cloud encoding. This enables the model to effectively assimilate comprehensive multi-modal attributes within the constraints of limited 3D data. Therefore, we fuse the 2D features with 3D object features to enhance object-level visual representation as follows:
\begin{equation}
    \mathbf{O} = MLP(\phi(F_{obj\_2D})+F_{obj\_3D}) + F_{obj\_3D}
\end{equation}
where $F_{obj\_2D}$ and $F_{obj\_3D}$ represent the 2D object features from the pre-trained model CLIP \cite{radford2021learning} and the 3D object features from the point cloud encoder, respectively. $F_{obj\_2D}$ is mapped by linear function $\phi$ to obtain fine-tuned features that have the same dimensions as $F_{obj\_3D}$. The MLP-based fusion of point cloud features with multi-modal 2D features may potentially obscure certain geometric information. To mitigate this issue, the enhanced features are subsequently integrated with the original point cloud features.

\subsection{Progressive Relational Learning}
\label{sec: b2n}
\begin{figure}[t]
  \centering
  \includegraphics[width=0.95\linewidth]{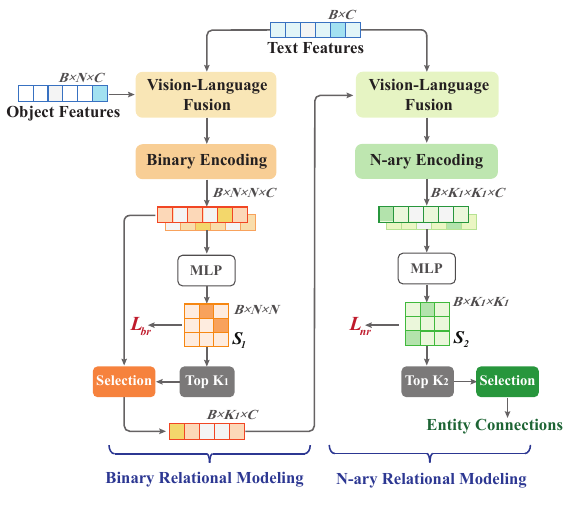}
  \caption{The structure of the \textbf{B2N-PRL} module. The inputs are object features $O_i \in \mathbb{R}^{B \times N \times C}$ and text features $T \in \mathbb{R}^{B \times C}$, where $B$ denotes the batch size and $N$ is the number of objects in one scene. Both features have the same dimensions, $C$. B2N-PRL sequentially models binary and n-ary relationships, ultimately outputting the $K_2$ entity combinations that best match the textual content.}
  \label{fig:b2n}
\vspace{-0.4cm}
\end{figure}

The textual descriptions for the target with distractors often involve multiple referential relationships, which are also seen as the combination of several pairwise object relationships. However, existing models predominantly employ pairwise relationship modeling followed by direct fusion with text modalities, exhibiting insufficient global contextual awareness in descriptions with n-ary relational dependencies. We reformulate the object grounding task as a multi-relational structure inference problem. The target entity is localized by identifying candidate object combinations that satisfy the composite relational constraints. We utilize the LLM to extract binary relationships with textual representation about entity names. The corresponding objects in the scene are identified using these entity names to supervise the relational learning model. We design the progressive relational learning module, B2N-PRL, for reasoning about binary to n-ary relationships, as shown in Fig. \ref{fig:b2n}.

\subsubsection{Soft Relational Label Generation}
\label{sec: LLM}
The LLM is used to acquire relational labels that correspond to binary combinations of entities derived from referential utterances. We provide a prompt like ``Find all of the binary relationships between entities in the input \textcolor{gray}{[description]}'' to extract all the binary relationships in the textual description related to the entities. For example, the input sentence is ``The rear pillow on the couch on the left hand side and below the shelves.'', the output of the LLM is “(pillow-couch, couch-shelves)”. These labels reflect the types of entities involved in referential relationships but do not specify individual object instances. Therefore, they can be interpreted as soft relational annotations for the uncertain representation of relationships between entities. The soft relational labels are subsequently employed to supervise the B2N-PRL module. Notably, LLM may generate non-entity combinations or other words that mismatch with object class names. To mitigate this problem, entity names that are inconsistent with objects are automatically re-matched to the closest names via LLMs, e.g. ``shelves'' to the canonical name ``shelf''. The semantic-aligned entity combinations derived from the outputs contain at least one configuration that precisely matches the spatial relationships in the textual inputs. We exploit this inherent property to construct supervisory signals for the model to guide the comprehension of binary and n-ary relationships.

\subsubsection{Binary Relational Modeling}
Binary relations as fundamental units for modeling complex spatial relationships are encoded from paired objects. We employ a BERT-based \cite{kenton2019bert} textual encoder to process natural language sentences and obtain text features $T$. During the perception process, sentence decomposition is unnecessary, and the holistic textual representation T directly engages in multi-modal interactions with visual information at various stages. Similarly to previous work \cite{chang2024mikasa}, we improve the spatial perception ability of the model by integrating box embedding $F_{box}$ and artificially extracted geometric features $F_{geo}$. The specific calculation is as follows:

\begin{equation}
\label{eq: 2}
    \mathbf{O'}_i=\mathbf{Att}_{cross}(\mathbf{O}_i+ F_{box_i},\mathbf{T})
\end{equation}
\vspace{-0.2cm}
\begin{equation} 
    \mathbf{B}_{i,j} = \mathbf{O'}_i\times \mathbf{O'}_j+F_{geo_ij}
\end{equation}

where $\mathbf{O}_i$ denotes the visual features of the $i$-th object, $\mathbf{Att}_{cross}$ represents the multi-head cross-attention operation, $\mathbf{B}_{i,j}$ denotes the binary relation vector of the $i$-th and $j$-th objects.

We map $\mathbf{B}$ through an MLP to the score map $S_1$ to represent the probability of all binary relationships matched with text. Since only the semantic categories of objects and targets are given in labels, the specific positions of the referred objects mentioned in descriptions are unknown. We assume that all binary relations combined with extracted entities in Section \ref{sec: LLM} should have higher probabilities, which is used to supervise the binary relation response value. The binary relational loss is defined as:
\begin{equation}
L_{br} = -\frac{1}{N \times N} \sum_{i=1}^{N \times N} \left[r_i \log(S_{1,i}) + (1 - r_i) \log(1 - (S_{1,i}) \right]
\end{equation}
where $r_i \in \{0,1\}$ indicates whether the $i$-th entity combination matches the description.

\subsubsection{N-ary Relational Modeling}
\label{sec: nre}
In complex scenes, the description typically pertains to not just one referred object. Instead, it usually necessitates the positional relationship with multiple referred objects to collectively define the target. We further incorporate the extracted binary relationship information to encode the n-ary relationships. Specifically, we select the top $K_1$ binary combinations from the response score in $S_1$, then obtain the representation of n-ary relationships as follows:
\begin{equation}
    \mathbf{B'}_i=\mathbf{Att}_{cross}(\mathbf{B}_i^{top-K_1},\mathbf{T})
\end{equation}
\begin{equation} 
    \mathbf{M}_{i,j} = \mathbf{B'}_i\times \mathbf{B'}_j\
\end{equation}
where $\mathbf{B}_i^{top-K_1}$ denotes the features of $i$-th binary relationship in the top $K_1$ combinations, the definitions of $\mathbf{Att}_{cross}$ and $T$ are same as in Equation (\ref{eq: 2}). Each n-ary relation vector $\mathbf{M}_{i,j}$ may involve 2 to 4 objects.

Similarly, we use an MLP to obtain the score map $S_2$ from $\mathbf{M}$ that represents the matching probability between n-ary relationships and the textual description. 
For training, we designed a new semi-supervised relational learning mechanism to divide the n-ary combinations into two groups: those with the target and those without it. The combinations that lack the target are likely inconsistent with the textual description, resulting in their probability of relevance to the text being nearly 0. For other n-ary combinations that contain the target, we assume that the combination with the highest probability value completely conforms to the text description, and its probability should be close to 1. The n-ary relational loss is defined as:
\begin{equation}
L_{nr} = -\frac{1}{N_{neg}} \sum_{i=1}^{N_{neg}} \log(1-S_{2.i}^{neg})-\log(max(S_2^{pos}))
\end{equation}
where $S_2^{pos}$ and $S_2^{neg}$represent the scores of two groups respectively, $N_{neg}$ is the number of n-ary combinations without the target. 

\subsection{Attention-Driven Graph Learning}
In order to further understand the spatial relationships of entities, we construct a graph attention network. This enables the model to localize the final target using information transmitted through multi-modal nodes. Existing methods based on constructing scene graphs for all entities or filtering out some nodes according to semantic priors rely entirely on the implicit learning of graph neural networks, which are still insufficient for 3D-language alignment of spatial relations. Our approach exploits text-related n-ary relations to specify the path of information transmission in scene graphs. The relational learning must only be performed on these entity combinations that respond more to text descriptions.

\subsubsection{Graph Construction via N-ary Relationships}
In Section \ref{sec: nre}, the consistency probabilities between each multi-tuple and the textual description are predicted as $S_2$. We select the top $K_2$ n-ary combinations from $S_2$ to construct the scene graph. All entities within n-ary combinations are considered as graph nodes. Each node is initialized with the input multi-modal features $O'$ from the binary relation module. Connections between nodes are established for all pairwise entities involved in the chosen $K_2$ n-ary relations. On one hand, our method globally identifies the most matching entity combination with the referential languages, using a progressive reasoning from binary to n-ary relations. On the other hand, each node interacts exclusively with those nodes that are potentially relevant in the described relations. Throughout this process, multi-modal information is a guiding mechanism, enhancing the perception of complex spatial relationships.

\begin{figure}[t]
  \centering
  \includegraphics[width=0.9\linewidth]{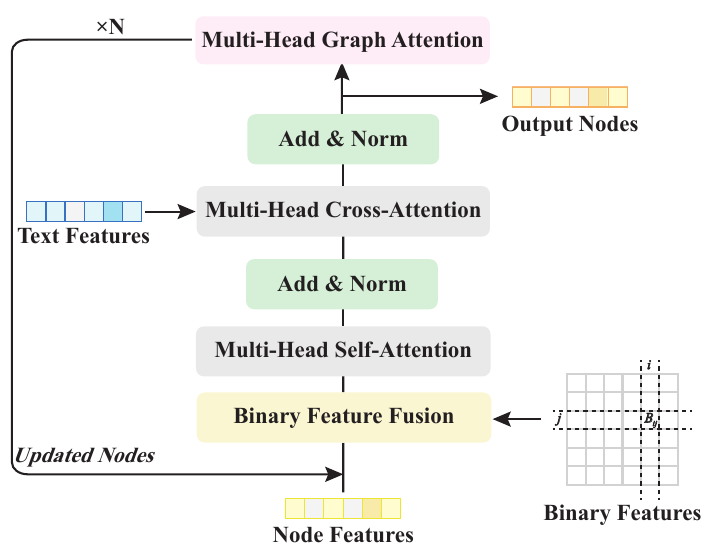}
  \caption{The graph-structured multi-modal network. The node features of the scene graph are input into the network. The graph node first fuses the binary features extracted from the B2N-PRL module and enhances the expression of spatial relationships through self-attention. Next, the text feature is fused with the node feature via cross-attention. The network contains $N$ graph attention layers to aggregate the node features for global relational perception.}
  \label{fig:gat}
\vspace{-0.4cm}
\end{figure}

\subsubsection{Graph-Structured Multi-Modal Network}
We integrate 3D vision with text in a graph structure and propose a multi-modal attention network. As shown in Fig. \ref{fig:gat}, we employ cross-attention to facilitate interaction between object features and text features, and graph attention to aggregate node information concerning referential relationships through n-ary connections. Specifically, node features are fused with binary features obtained from the relation learning module using a self-attention mechanism to capture object-to-object dependencies. Then we utilize the multi-head cross-attention operation to fuse the textual information and achieve global perception within a graph attention layer. The graph attention operation updates the node values as follows:
\begin{equation}
v_i^{'} = \rVert_{k=1}^{K}\sigma (\sum_{j\in E_i} \alpha_{ij}^{k} W^{k} v_j)
\end{equation}
where $v_i$ represents the feature vector of the $i$-th node, $E_i$ is its neighboring set, $v_j$ denotes the node feature of the $j$-th neighbor, $\alpha_{ij}^{k}$ is the attentive weight of the $k$-th head, $W^{k}$ is the projection matrix, and $\sigma()$ indicates a nonlinear activation layer. The graph attention aggregation process is performed two times throughout the entire network, resulting in the output of the updated node information.

\subsection{Loss Function}
During training, we set up the localization loss $L_{ref}$ to supervise the final target prediction results and use the classification loss $L_{t}$ and $L_{v}$ to guide the textual and visual encoding as in previous work \cite{chen2022multi,guo2023viewrefer,chang2024mikasa,zhang2024cross}. It is noted that previous methods use original instance classes as the gt labels for $L_{v}$. We reduce the number of mapped classes to rely less on the classification information inherent in the visual modality. In addition to them, we propose binary and n-ary relational losses ($L_{br}$ and $L_{nr}$ in Section \ref{sec: b2n}) to supervise the B2N-PRL module. The whole loss function is grouped as:
\begin{equation}
Loss = L_{ref}+\lambda_1 L_{t}+\lambda_2L_{v}+\lambda_3(L_{br}+L_{nr})
\end{equation}

where $L_{ref}$, $L_{t}$, and $L_{v}$ are computed using the cross-entropy loss function,  $\lambda_1-\lambda_3$ are weight coefficients.

\section{Experiment}

\subsection{Datasets and Metrics}
Our method is evaluated on three 3D object grounding datasets: Nr3D, Sr3D, and ScanRefer. Nr3D and ScanRefer contain 41,503 and 51,583 human-annotated descriptions, respectively, while Sr3D includes 83,572 descriptions synthesized from five pre-set spatial relation templates. All objects and scenes described in these datasets are sourced from the large-scale 3D indoor dataset ScanNet, with the targets primarily located by the referential relationships among entities.

The evaluation of Nr3D and Sr3D follows the guidelines set by the Referit3D benchmark, where all text-based localizations rely on ground-truth 3D bounding boxes. Referit3D presents two main challenges: firstly, there are inevitably distractors within the same semantic category as the target in the scene; secondly, some positional descriptions depend on the observed viewpoints indicated in the utterances. ScanRefer conducts localization directly on the original scene point clouds, and its metrics require calculating accuracy based on the Intersection over Union (IoU) of the predicted and actual target boxes.

\subsection{Implementation Details}
We use GPT-4o mini \cite{achiam2023gpt} as the LLM-based agent to extract entity combinations related to spatial relationships from descriptions. All multi-head cross-attention, self-attention, and graph attention layers in our methods are configured with eight heads. The loss weight coefficients $\lambda_1$, $\lambda_2$, and $\lambda_3$ are set to 0.1, 0.5, and 2.0, respectively. The models are trained with a batch size of 20 for 150 epochs, using the Adam \cite{kingma2014adam} optimizer. The learning rate is initialized to $5 \times 10^{-4}$ with a decay of 0.65 every 10 epochs from 30 to 80. In the B2N-PRL module, the number of selected binary and n-ary relationships is set to $K_1$=16 and $K_2$=16.

\begin{table*}[!t]
\begin{center}
	\caption{Grounding accuracy (\%) on Nr3D and Sr3D datasets with ground-truth object proposals.}
    \renewcommand\arraystretch{1.3}
    \fontsize{9pt}{10pt}\selectfont
    \tabcolsep=6pt
    \label{tab1}
    \begin{threeparttable}
	\begin{tabular}{c|c|ccccc|ccccc}
    \Xhline{2pt}
	  \multirow{2}{*}{\textbf{Methods}}&\multirow{2}{*}{\textbf{Year}}&\multicolumn{5}{c|}{\textbf{Nr3D}} &\multicolumn{5}{c}{\textbf{Sr3D}}\\
        \cline{3-7}\cline{8-12}
		&  &Overall  &Hard &Easy   &V-dep  &V-indep &Overall    &Hard &Easy  &V-dep  &V-indep \\
       \cline{1-12}
		ReferIt3D \cite{achlioptas2020referit3d} &\textit{2020}  &35.6 &27.9&43.6   &32.5 &37.1 &40.8 &31.5 &44.7 & 39.2 &40.8 \\
	  TGNN \cite{huang2021text} &\textit{2021}  &37.3 &30.6&44.2  &35.8 &38.0 &45.0 &36.9&48.5  & 45.8 &45.0 \\     
        3DVG-Trans \cite{zhao20213dvg} &\textit{2021} &40.8  &34.8&48.5  &34.8 &43.7 &51.4 &44.9&54.2  &44.6 &51.7 \\
        FFL-3DOG \cite{feng2021free} &\textit{2021} &41.7 &35.0&48.2   &37.1 &44.7 &- &- &- &- &- \\
        LAR \cite{bakr2022look} &\textit{2022} &48.9 &42.3&58.4   &47.4 &52.1 &59.4  &51.2 &63.0&50.0 &59.1 \\
        SAT \cite{yang2021sat}&\textit{2021}   &49.2  &42.4&56.3  &46.9 &50.4 &57.9 &50.0&61.2  &49.2 &58.3 \\   
        M3DRef-CLIP \cite{zhang2023multi3drefer} &\textit{2023} &49.4 &43.4&55.6   &42.3 &52.9 &- &- &- & - &- \\
        3D-SPS \cite{luo20223d} &\textit{2022} &51.5 &45.1&58.1   &48.0 &53.2 &62.6 &65.4&56.2  & 49.2 &63.2 \\
        EDA \cite{wu2023eda} &\textit{2023} &52.1 &-  &- &- &- &68.1 &- &- &- &- \\
        BUTD-DETR \cite{jain2022bottom}&\textit{2022}   &54.6  &48.4&60.7  &46.0 &58.0 &67.0 &63.2&68.6  &53.0 &67.6 \\         
        MVT \cite{huang2022multi}&\textit{2022}  &55.1  &49.1&61.3  &54.3 &55.4 &64.5 &58.8&66.9  &58.4 &64.7 \\
        ViewRefer \cite{guo2023viewrefer}&\textit{2023}  &56.0  &49.7&63.0  &55.1 &56.8 &67.0 &62.1&68.9  &52.2 &67.7 \\   
        ViL3DRel \cite{chen2022language}  &\textit{2022} &64.4 &57.4&70.2  &62.0 &64.5 &72.8 &67.9 &74.9 &63.8 &73.2 \\
        MiKASA \cite{chang2024mikasa} &\textit{2024} &64.4  &59.4&69.7  &65.4 &64.0 &75.2 &67.3&78.6  &\textbf{70.4} &75.4 \\          
        MA2TransVG \cite{xu2024multi} &\textit{2024} &65.2 &57.6&71.1   &62.5 &65.4 &73.9 &69.3&76.0 & 64.5 &73.8 \\ 
        xM\_Match \cite{zhang2024cross} &\textit{2024} &66.2 &59.9 &72.8   &63.8 &67.5 &74.6 &\textbf{71.3}&75.9  &65.0 &74.7 \\ 
        \hline    							 
        \textbf{B2N3D (ours)}  &- &\textbf{68.3}  &\textbf{62.7} &\textbf{74.1} &\textbf{67.8} &\textbf{68.6} &\textbf{76.6} &70.1 &\textbf{79.4} &66.5 &\textbf{77.1} \\ 				
    \Xhline{2pt}   				 				
	\end{tabular}
 \end{threeparttable}
\end{center}
\vspace{-0.3cm}
\end{table*}

\begin{table}[!t]
\begin{center}
 \caption{Grounding accuracy (\%) on ScanRefer with detected object proposals}
    \renewcommand\arraystretch{1.2}
    \fontsize{9pt}{10pt}\selectfont
    \tabcolsep=2pt
    \label{tab2}
     \begin{threeparttable}
	\begin{tabular}{ccccc}
    \Xhline{2pt}
	  Methods&Category&\hspace{0.5em} 2D Encoder\hspace{0.5em} &Multiple&Overall\\
        \hline
        3D-SPS \cite{luo20223d} &one-stage &Pixel-N &29.8 &37.0\\
        BUTD-DETR \cite{jain2022bottom} &one-stage &- &32.8 &37.1\\
        3DVLP \cite{zhang2024vision} &one-stage &Pixel-N  &33.4  &40.5\\  
        EDA \cite{wu2023eda} &one-stage &- &37.6 &42.3\\
        MCLN \cite{qian2024multi}  &one-stage&-  &38.4  &42.6\\
        \hline 
        TGNN \cite{huang2021text} &two-stage & &23.2 &29.7\\
        InstanceRefer \cite{yuan2021instancerefer} &two-stage &- &24.8 &32.9\\
        ViewRefer \cite{guo2023viewrefer} &two-stage &- &26.5 &33.7\\
        ViL3DRel \cite{chen2022language} &two-stage &- &30.7 &37.7\\
        3D-LP \cite{jin2023context} &two-stage& &32.2 &38.1\\
        xM\_Match \cite{zhang2024cross}  &two-stage &- &- &39.3\\
        \textbf{B2N3D (ours)} &two-stage  &- &\textbf{38.1} &\textbf{42.7}\\ 
       \hline      
        ScanRefer \cite{chen2020scanrefer} &two-stage &Pixel-N &21.1 &27.4\\
        3DVG-Trans \cite{zhao20213dvg} &two-stage  &Pixel-N&28.4 &34.7 \\ 
        3DJCG \cite{cai20223djcg} &two-stage &Pixel-N &30.8 &37.3\\
        D3Net \cite{chen2022d}&two-stage &Pixel-N  &30.1 &37.9\\
        3D-LP \cite{jin2023context}&two-stage &Pixel-N &33.4 &39.5\\
        MA2TransVG \cite{xu2024multi}  &two-stage &Pixel-N &41.4  & 45.7\\ 
        \hline
        M3DRef-CLIP \cite{zhang2023multi3drefer} &two-stage &Object-3 &36.8  &44.7\\ 
        \textbf{B2N3D (ours)} &two-stage &Object-1 &\textbf{41.1} &\textbf{45.6} \\   			
    \Xhline{2pt}
	\end{tabular}
 \end{threeparttable}
\end{center}
\vspace{-0.4cm}
\end{table}

\begin{table}[!t]
\begin{center}
	\caption{Ablation studies of key components on Nr3D and ScanRefer}
    \renewcommand\arraystretch{1.3}
    \fontsize{9pt}{10pt}\selectfont
    \tabcolsep=5pt
    \label{tab3}
     \begin{threeparttable}
	\begin{tabular}{cc|c|c|c}
    \Xhline{2pt}
        \multicolumn{2}{c|}{B2N-PRL}&\multirow{2}{*}{Scene Graph}&\multirow{2}{*}{Nr3D}&\multirow{2}{*}{ScanRefer}\\
         \cline{1-2}
	   Binary & N-ary & & &\\
         \hline 					
         \checkmark&\checkmark& n-ary combinations &\textbf{68.3} &\textbf{45.6}\\ 
        \checkmark&- & binary combinations &67.3 &45.1\\ 
           -&- & full-connected &66.6 &44.8\\  			
           -&- & - &65.9 &44.4\\  	
    \Xhline{2pt} 
	\end{tabular} 
 \end{threeparttable}
\end{center}
\vspace{-0.4cm}
\end{table}

\begin{table}[!t]
\begin{center}
	\caption{Ablation studies with various $K_1$ and $K_2$ settings on Nr3D}
    \renewcommand\arraystretch{1.3}
    \fontsize{9pt}{10pt}\selectfont
    \tabcolsep=5pt
    \label{tab5}
	\begin{tabular}{cc|ccccc}
    \Xhline{2pt}
         $K_1$&$K_2$ & Overall &Hard &Easy &V-dep  &V-indep\\
        \cline{1-7}   		
           8&8 &65.5 &59.1 &72.2 &64.2 &66.1 \\ 								
           16& 8 &67.4  &61.8 &73.3  &65.9 &68.2\\  								
           \rowcolor{gray!20} 16& 16 &\textbf{68.3}  &\textbf{62.7} &\textbf{74.1} &\textbf{67.8} &\textbf{68.6} \\ 	
           32&16 &65.9 &59.8 &72.3 &65.5 &66.1\\ 					
           32&32 &66.0  &59.5  &72.6 &65.0 &66.4\\   				
    \Xhline{2pt}
	\end{tabular}
\end{center}
\vspace{-0.5cm}
\end{table}

\begin{figure*}[!t]
  \centering
  \includegraphics[width=0.95\linewidth]{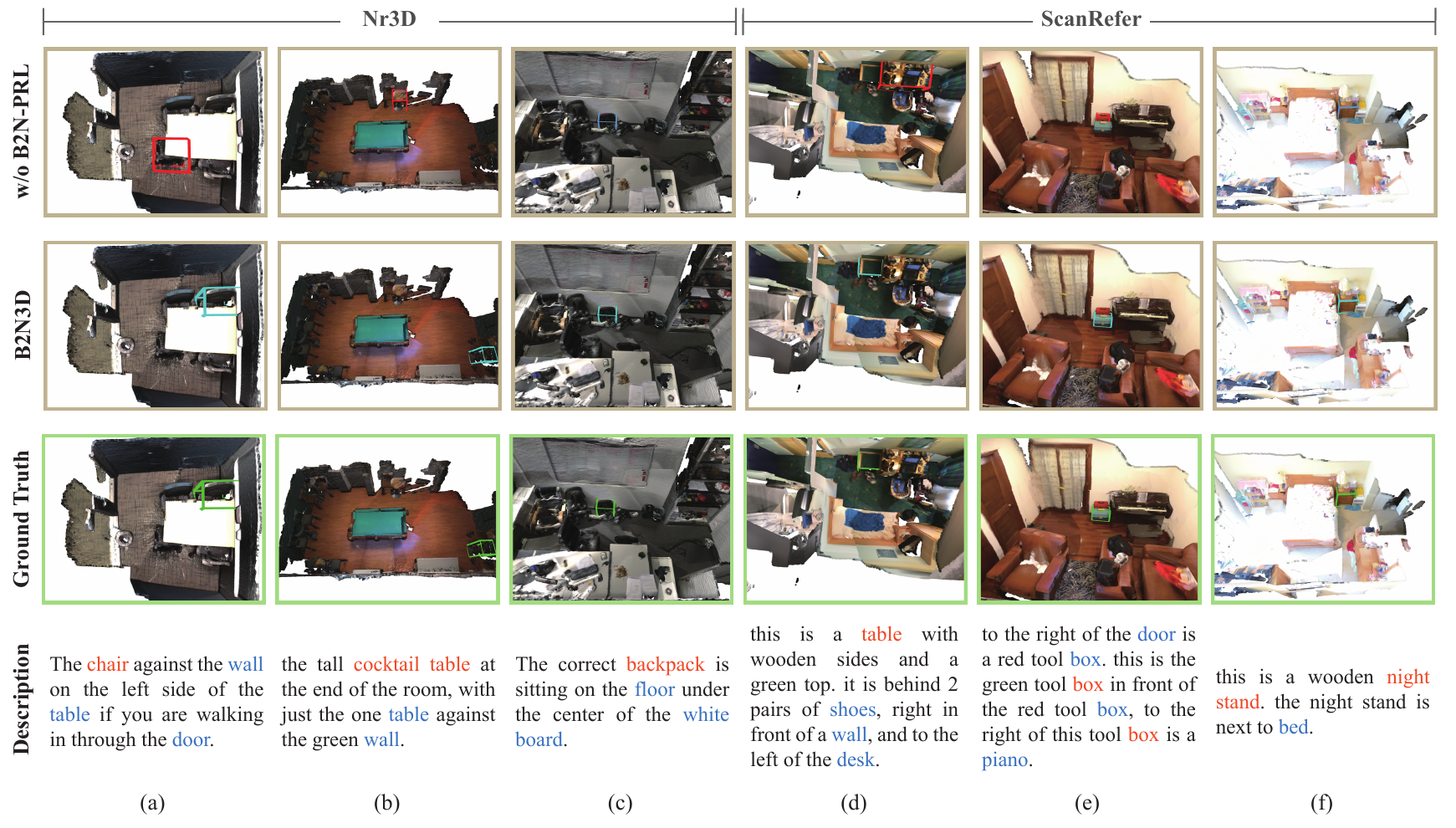}
  \caption{The visualization examples of B2N3D on the human-annotated datasets Nr3D and ScanRefer. The first row is the result of the model without progressive relational learning. 3D bounding boxes represent the grounding results and ground truth. The entity words in the description are marked in orange for the target and in blue for the referred objects.}
  \label{fig:vis}
  \vspace{-0.5cm}
\end{figure*}

\subsection{Comparison with State-of-the-Art Methods}
To evaluate the effectiveness of our proposed B2N3D, we compare it with state-of-the-art approaches on Nr3D, Sr3D, and ScanRefer datasets. As shown in Table \ref{tab1}, our method achieves superior performance in overall localization accuracy. Specifically, B2N3D achieves 62.7\% on hard samples and 67.8\% on view-dependent samples of Nr3D, which are significantly higher than the latest method xM\_Match, demonstrating our advantage in challenging cross-modal understanding. The synthetic dataset, Sr3D, has a lower proportion of complex relation descriptions (Fig. \ref{fig:llm}). Thus, the enhancements brought about by our progressive learning method are subtly noticeable. 

Table \ref{tab2} shows the evaluation results on ScanRefer, where ``Multiple'' is similar to ``Hard'' of Nr3D, indicating that the targets share the same-category distractors. In the scenario where only 3D point cloud data is utilized, our model outperforms other approaches, surpassing the latest two-stage method xM\_Match by 3.4\%. Many methods enhance the color and texture features of point clouds by combining 2D image information, leading to more accurate localization. However, most of these encode all multi-view information related to point clouds at the pixel level (``Pixel-N'' in Table \ref{tab2}), resulting in high computational costs. In contrast, object-level fusion typically requires only a few views and is integrated with the point cloud network. Compared to M3DRef-CLIP, which uses three object views, our model requires only a single object view and accomplishes a markedly improved localization accuracy.

In addition, the qualitative results in Fig. \ref{fig:vis} further validate the effectiveness of our method in precisely localizing objects defined in natural language descriptions. This is particularly notable in complex scenarios where the target location hinges on multiple referential relations. Overall, these results demonstrate that our method not only achieves competitive performance in 3D object grounding but also offers global perception advantages in terms of complex relationships.

\begin{table*}[!t]
\begin{center}
   \caption{Examples of soft relational labels produced by the LLM.}
    \renewcommand\arraystretch{1.3}
    \fontsize{9pt}{10pt}\selectfont
    \label{tab4}
   \begin{threeparttable}
	\begin{tabular}{c|p{8cm}|p{7cm}|c}
      \Xhline{2pt} 
        &Description&Output&RN\\
         \hline
          \rowcolor{gray!10} 1&the larger of the two pillows on the couch in front of the other pillow& pillow-couch, pillow-pillow &2 \\
           2&the desk with one chair in front of it& desk-chair &1\\
          \rowcolor{gray!10} 3&the smallest kitchen cabinet on the floor closest to the trash can and door &kitchen cabinet-door, kitchen cabinet-trash can, kitchen cabinet-floor & 3\\
          4&there are two pillows next to the fridge, pick the one that is on the edge of the couch &pillow-couch, pillows-refrigerator& 2\\
          \rowcolor{gray!10} 5&this stool is underneath the table away from the paintings, and is on the left side from the perspective of looking towards the paintings &stool-table, table-picture, stool-picture&3\\
          \hline
          6&looking at the stack of pillows, please select the bottom pillow&chair-pillow, pillow-stack of chairs & error \\
          \rowcolor{gray!10}7&the desk is in the middle of the room, further from the window  to the right of the entrance&desk-door, desk-bed,desk-window& error \\
          8& pick the box that is sitting on the desk against the wall with shelves and a white board &desk-wall, box-desk, desk-shelf, desk-blackboard &error\\							
       \Xhline{2pt}   
	\end{tabular} 
 \end{threeparttable}
\end{center}
\end{table*} 

\begin{figure*}[!t]
  \centering
  \includegraphics[width=0.95\linewidth]{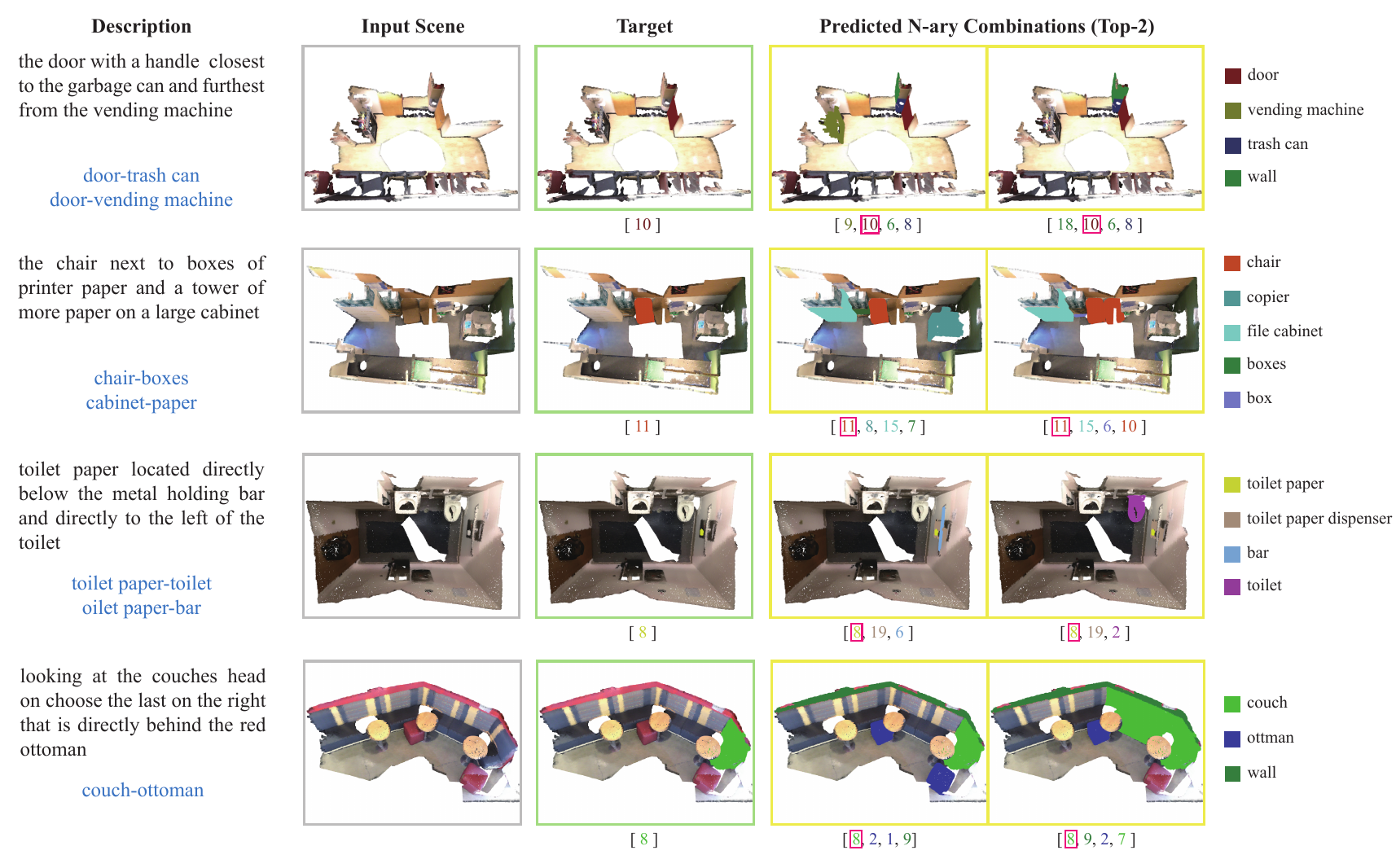}
  \caption{The n-ary entity combinations predicted to have the highest response by the B2N-PRL module are visualized in the point cloud using 3D masks. The numeric values enclosed in ``[]'' indicate the corresponding object IDs within the scene. The entity word combinations derived from the textual description appear in blue font. Notably, objects within an n-ary combination can simultaneously participate in multiple entity relations, thus fulfilling the requirement for global perception.}
  \label{fig:n_ary}
  \vspace{-0.5cm}
\end{figure*}

\subsection{Ablation Studies}

\subsubsection{Components of Relational Learning}
Our proposed B2N-PRL module consists of two core components: binary relational modeling and n-ary relational modeling. The model progressively infers from binary to n-ary relationships, thereby effectively capturing object information that satisfies multiple spatial constraints. Table \ref{tab3} displays the performance when only binary relationships are modeled, as well as the results obtained without relational reasoning. In our original framework, the scene graph is constructed based on predicted n-ary entity combinations. To evaluate the contribution of the B2N-PRL module, we conducted two comparative experiments using a fully connected scene graph and one that excludes scene graph learning entirely (corresponding to the last two rows). As shown in Table \ref{tab3}, the B2N-PRL module significantly improves scene graph construction and enhances multi-modal global perception.

 As (a) in Fig. \ref{fig:vis}, it is necessary to combine the multiple positional relationships between ``chair'' and ``wall'', ``table'', and ``door'' to select the correct ``chair'' from the four. Our proposed B2N-PRL module enables precise localization of the target through global perception. For targets that can be identified based on simple contextual cues, such as the ``backpack'' in (c), where its spatial relationship with either the ``floor'' or the ``whiteboard'' is sufficient for localization, correct results can still be achieved without global relational learning.

\subsubsection{Hyperparameters of Relational Modeling}
In our progressive learning framework, the top $K_1$ or $K_2$ entity combinations with the highest probabilities at each level are utilized for relational perception at the next level. The performance comparison of different numbers of $K_1$ and $K_2$ on the Nr3D dataset is shown in Table \ref{tab4}. The evaluation on Nr3D is based on ground-truth proposals, which eliminates the impact of detection differences on the effectiveness of relational learning. It can be observed that the overall performance is optimal when both $K_1$ and $K_2$ are set to 16.

\begin{figure*}[t]
  \centering
  \includegraphics[width=0.8\linewidth]{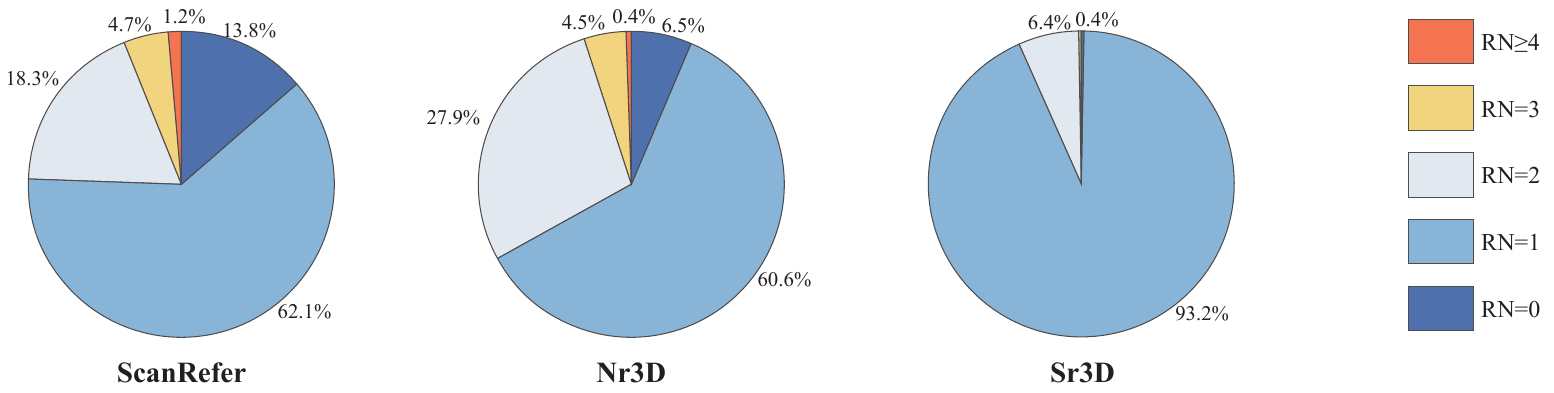}
  \caption{The pie charts present a statistical overview of the entity relationships in the training sets of Nr3D, Sr3D, and ScanRefer, where the relationships are automatically extracted using the LLM. ``RN''  denotes the number of binary entity combinations.}
  \label{fig:llm}
\vspace{-0.4cm}
\end{figure*}

\subsection{Discussion about Soft Relational Labels}
To enable the model to progressively capture n-ary relational semantics, we create a soft relational label generation mechanism for training. Specifically, the LLM is used to extract pairwise entity relations from textual descriptions. Some representative examples involving different numbers of relationships (``RN'') are presented in Table \ref{tab4}, along with instances of incorrect generations. It can be seen that the LLM may hallucinate entities that are not explicitly mentioned in the input text. For example, the words ``chair'' in 6 and ``bed'' in 7 do not appear in the original sentences, and ``white board'' in 8 is incorrectly recorded as "blackboard". Such hallucinations may distort the relational structure and introduce noise into the extracted relational labels, affecting the training process of binary relational modeling. In our designed n-ary relational supervision framework, the uncertainty of soft labels is explicitly considered. The model learns global cross-modal correspondences in a target-oriented way, which helps reduce the negative effects of noisy labels.

Fig. \ref{fig:llm} illustrates the perception of relationship numbers in soft labels across various datasets. Both Nr3D and ScanRefer consist of human-annotated descriptions with more complex referential expressions, resulting in a higher number of binary relationships involving more than two entities. In contrast, Sr3D is a template-based synthetic dataset, where single spatial relations dominate. As shown in the statistical results, multi-relational descriptions occupy a considerable proportion in the datasets, indicating that learning from single binary relationships alone is insufficient for understanding complex textual inputs. Experimental results shown in Table \ref{tab5} demonstrate that our proposed B2N-PRL module significantly improves localization performance for understanding two or more referential relationships. The largest gain is seen on the Nr3D dataset, where accuracy increases from 60.6\% to 67.7\%. 

\begin{table}[!t]
\begin{center}
	\caption{Performance of various relational modeling methods involving more than two relationships}
    \renewcommand\arraystretch{1.3}
    \fontsize{9pt}{10pt}\selectfont
    \tabcolsep=3pt
    \label{tab5}
   \begin{threeparttable}
	\begin{tabular}{c|ccc|ccc}
      \Xhline{2pt} 
        &\multicolumn{3}{c|}{Nr3D}&\multicolumn{3}{c}{ScanRefer}\\
        \cline{2-7}
	 &Overall&RN$\geq$2 &RN$\leq$1 &Overall&RN$\geq$2 &RN$\leq$1\\
        \hline 				
        B2N3D &68.3 &67.7 &68.3&45.6 &50.6 &44.1 \\ 	
        w/o B2N-PRL &66.6 &60.6 &66.7 &44.8 &49.2 &43.5\\      				
       \Xhline{2pt}   
	\end{tabular} 
 \end{threeparttable}
\end{center}
\vspace{-0.4cm}
\end{table}

\section{Conclusion}
We propose a novel 3D object grounding framework that can improve the cross-modal understanding of multi-relation descriptions. Our method achieves global perception of relational conditions through progressive learning from binary to n-ary relationships and attention-driven graph learning, thereby enabling more accurate object localization. We creatively utilize the semantic understanding capabilities of LLMs to analyze entity relationships from natural languages and design a soft-label supervision mechanism to train the relation learning module. We experimentally validate the effectiveness of our method on three public datasets, achieving significant improvements over existing methods. For scenes containing a large number of objects, the computational cost of relation modeling becomes relatively high. In future work, we will focus on learnable conditional modeling mechanisms to reduce the computational overhead.

\bibliographystyle{IEEEtran}
\bibliography{literature}

\end{document}